\documentclass[a4paper,conference]{IEEEtran}
\usepackage{times}
\usepackage{epsfig}
\usepackage{graphicx}
\usepackage{amsmath}
\usepackage{amssymb}
\usepackage{caption}
\usepackage[nocompress]{cite}
\usepackage{slashbox}
\usepackage[dvipsnames,table]{xcolor} 
\usepackage{enumerate}
\usepackage{tabularx,ragged2e}
\usepackage{spreadtab}
\usepackage{adjustbox}
\usepackage{booktabs}
\usepackage{xcolor}
\usepackage{multirow}
\usepackage{rotating}
\usepackage{color, colortbl}
\usepackage{subfigure}

\usepackage[inline]{enumitem}

\usepackage[pagebackref=true,breaklinks=true,letterpaper=true,colorlinks,bookmarks=false]{hyperref}

\graphicspath{{./figures/}}

\DeclareGraphicsExtensions{.jpeg,.png,.eps,.pdf}

\definecolor{Gray}{gray}{0.9}

\definecolor{Gray}{gray}{0.85}
\definecolor{LightCyan}{rgb}{0.88,1,1}

\newcolumntype{a}{>{\columncolor{Gray}}c}
\newcolumntype{b}{>{\columncolor{white}}c}

\newcommand{\ptspace}{\vspace*{5pt}}
\newcommand{\partitle}[1]{\bigbreak\noindent\textbf{#1}}

\makeatletter
\newcommand*{\rom}[1]{\expandafter\@slowromancap\romannumeral #1@}
\makeatother

\newcommand{\Ree}{\mathbb{R}}

\newcommand{\etal}{\textit{et al}. }
\newcommand{\ie}{\textit{i}.\textit{e}., }
\newcommand{\eg}{\textit{e}.\textit{g}. }

\definecolor{applegreen}{rgb}{0.55,0.71,0.0}
\definecolor{babyblue}{rgb}{0.54,0.81,0.94}
\definecolor{azure}{rgb}{0.0,0.5,1.0}
\definecolor{budgreen}{rgb}{0.48,0.71,0.38}
\definecolor{amaranthpurple}{rgb}{0.67,0.15,0.31}

\ifCLASSINFOpdf
\else
\fi
\begin{document}
%
\title{Self-Selective Context for Interaction Recognition}

\author{\IEEEauthorblockN{Mert Kilickaya,
Noureldien Hussein,
Efstratios Gavves, Arnold Smeulders}
\IEEEauthorblockA{QUvA Deep Vision Lab,
University of Amsterdam\\
Netherlands\\
Email: kilickayamert@gmail.com, \{nhussein, egavves, a.w.m.smeulders\}@uva.nl}}


%


\maketitle

\begin{abstract}


Human-object interaction recognition aims for identifying the relationship between a human subject and an object. Researchers incorporate global scene context into the early layers of deep Convolutional Neural Networks as a solution. They report a significant increase in the performance since generally interactions are correlated with the scene (\ie riding bicycle on the city street). However, this approach leads to the following problems. It increases the network size in the early layers, therefore not efficient. It leads to noisy filter responses when the scene is irrelevant, therefore not accurate. 
It only leverages scene context whereas human-object interactions offer a multitude of contexts, therefore incomplete. To circumvent these issues, in this work, we propose Self-Selective Context (SSC). SSC operates on the joint appearance of human-objects and context to bring the most discriminative context(s) into play for recognition. We devise novel contextual features that model the locality of human-object interactions and show that SSC can seamlessly integrate with the State-of-the-art interaction recognition models. Our experiments show that SSC leads to an important increase in interaction recognition performance, while using much fewer parameters.  


\end{abstract}


%
\IEEEpeerreviewmaketitle

\section{Introduction}

The goal of this paper is to recognize Human-object interactions from a single image. Human-object interaction recognition is an important problem with applications in areas such as robotics~\cite{lallee2010human,yu2018one,edsinger2007human}, image captioning~\cite{herdade2019image} or visual question answering~\cite{mallya2016learning}. The task is to identify the relationship between a human and an object in terms of a \texttt{<verb, noun>} pair, such as \texttt{<ride, bicycle>}. Since the type of interaction is highly correlated with the scene (\ie riding bicycle on the city street), researchers tackle this problem via integrating scene into the deep Convolutional Neural Networks (CNNs)~\cite{gkioxari2015actions,mallya2016learning,intrec-partattention}.

Initially, Gkioxari~\etal~\cite{gkioxari2015actions} augments human appearance with the global scene in a late-fusion manner by combining human and scene classifiers. However, late-fusion cannot model the correlation between the human and the scene. To that end, Mallya and Lazebnik~\cite{mallya2016learning} combines human appearance with the global scene early in the network layers, leading to a significant increase in the performance. Later, Fang~\etal~\cite{intrec-partattention} improves this model by further augmenting early-fused global scene appearance with the scene objects. 



\begin{figure}[t]
\begin{center}
\includegraphics[width=0.75\linewidth]{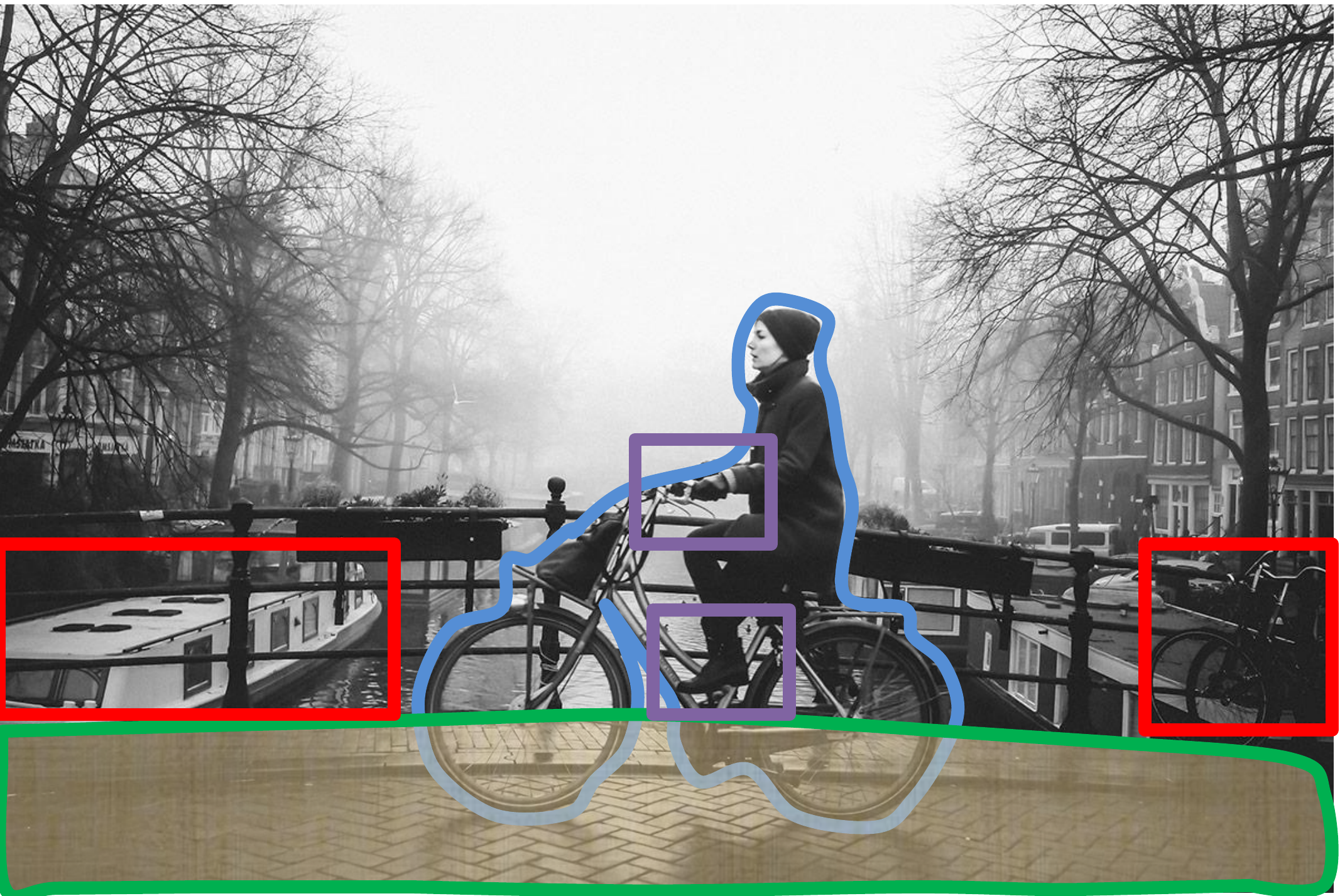}
\end{center}



\caption{Human-object interactions come with many contexts that can help in recognition. In the example above, utilizing the \textcolor{amaranthpurple}{body-parts}, the \textcolor{azure}{deformation}, and the \textcolor{budgreen}{surround} scene can ease the recognition of \texttt{<ride, bicycle>}. However, background \textcolor{red}{objects} like the boats can mislead the recogniton. In this paper, we first develop novel contextual features, as well as a context selection scheme Self-Selective Context to rely only on the most discriminative contexts.}

\label{fig:teaser}
\end{figure}

We identify the following problems with this approach. First, incorporating scene early in the network layers increases network parameters, limiting the efficiency. Second, changes in the scene appearance (\ie a clutter object) yields noisy filter responses, limiting the accuracy. Third, Human-object interactions offer a multitude of contexts beyond the global scene, see Figure~\ref{fig:teaser}, which is yet to be explored.



In this work, we propose Self-Selective Context (SSC) to circumvent the aforementioned problems. SSC learns to select the discriminative context(s) conditioned on the input image. It considers the joint appearance of human-object and context to decide which context feature(s) are discriminative. To take advantage of the multitude of contextual features offered by human-object interactions, we also devise contextual features that model the locality of the interactions. Our experiments on three large-scale benchmarks reveal that indeed the proposed contextual features are discriminative, and SSC can further boost the performance by selecting the discriminative feature in a scalable fashion.  

Our contributions are as follows: 

\begin{enumerate}
    \item We propose novel contextual features to represent the locality of humans, objects, scene, and human-objects.
    \item We propose Self-Selective Context to selectively utilize the discriminative context depending on the input image. 
    \item On three benchmarks for interaction recognition, SSC, combined with our novel contextual features, improves baseline models while being three times more parameter-efficient. 
\end{enumerate}



\begin{figure*}[!ht]
\begin{center}
\includegraphics[trim=0mm 8mm 0mm 0mm,width=0.75\textwidth]{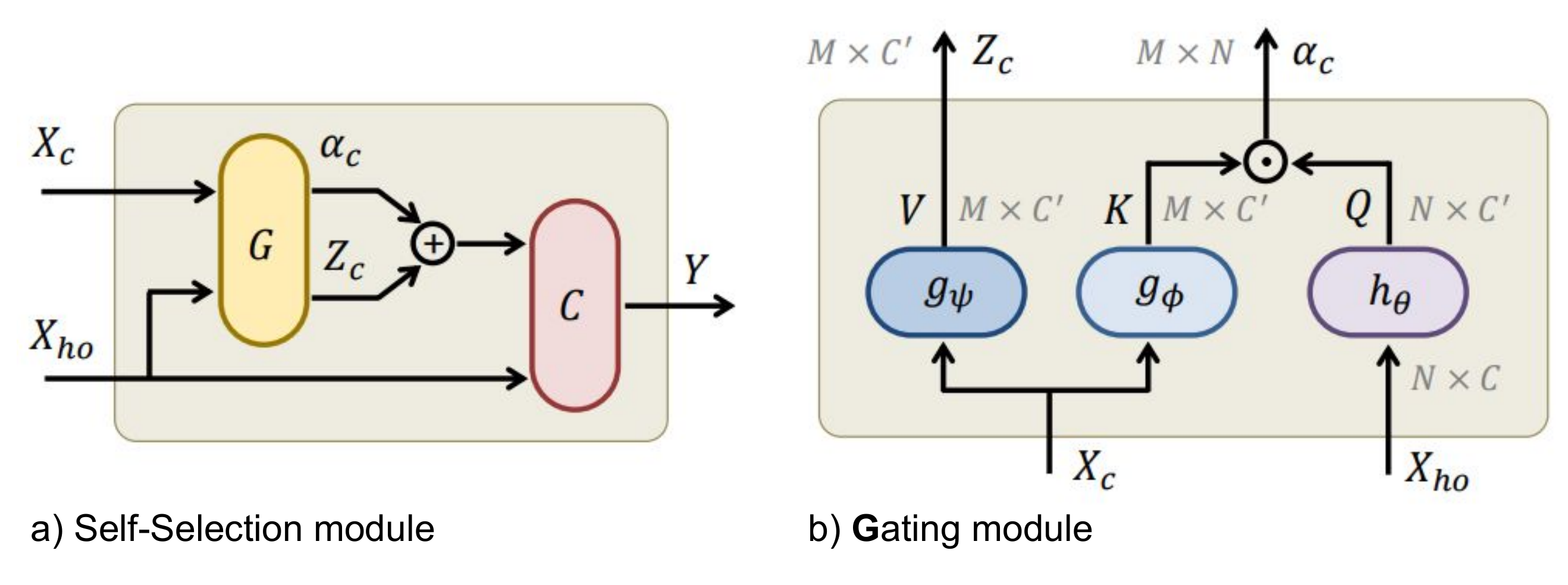}
\end{center}
\caption{
Overview of our method.
On the left, the Self-Selection module.
It takes as an input the features $\mathbf{X}_{ho}$ of $N$ human-object pairs in a certain image, and $M$ context features $\mathbf{X}_c$ corresponding to the image.
Then, it modulates the context features $\mathbf{X}_c$ using a novel Gating module $G(\cdot)$.
The final image-level features are then feed-forwarded to the classifier $C(\cdot)$ to predict the human-object interactions in the image.
On the right, the Gating module $G(\cdot)$, inspired by the Self-attention~\cite{vaswani2017attention}.
The main purpose of $G(\cdot)$ is to embed the heterogeneous context features $\mathbf{X}_c$ into a compact representation $\mathbf{Z}_c$.
$G(\cdot)$ predicts the vectors $\alpha$ used for Self-Selection of the embedded context features $\mathbf{Z}_c$.}
\label{fig:gating}
\end{figure*}

\section{Related Work}

\subsection{Human-object interaction recognition} 


Recently there has been good progress on interaction detection~\cite{chao2018learning,gkioxari2018detecting} which requires bounding box annotations for each interaction in the image. In our work, we instead focus on interaction recognition, which is an image-level classification task since many images exhibit humans manipulating objects.


Human-object interaction recognition is defined as single image multi-label classification task in~\cite{hico}. The authors collect a large-scale dataset named HICO with multiple image-level annotations for concurrent interactions, such as \texttt{<ride, bicycle>} and \texttt{<hold, bicycle>}. In our paper, we resort to the definition of HICO for interaction recognition. HICO allows researchers to train deep CNNs where they combine human and global scene context~\cite{gkioxari2015actions,mallya2016learning,intrec-partattention} to classify the interaction. 

HICO dataset is collected by \texttt{<verb, noun>} queries-only, therefore the contextual diversity is limited (\ie most interactions occur in their canonical contexts). 
This prevents seeing the models generalization abilities across different environments of the same interaction. 
To tackle this, in our paper we collect a new dataset we name \textbf{C}ontextualized \textbf{Int}eractions (CINT) which exhibits interactions within diverse contexts. In addition, we develop novel context features to leverage the locality of the human-object interactions. The locality is more robust to changes in the visual context, as we demonstrate through our experiments on HICO~\cite{hico}, V-COCO~\cite{vcoco}, and CINT.

\subsection{Combining multiple contexts in action recognition} 
Multiple cues are helpful in action recognition, where the dominant approaches are fusion~\cite{fusionbagwords,karpathy2014large,fusionmotion}. Such approaches can handle single context, however, they fall short in case of multiple contexts. Early fusion scales quadratically with the number of fused contexts. Late fusion does not leverage the correlation of human-object feature and context feature. To that end, in our work, as inspired by the Self-attention~\cite{vaswani2017attention}, we develop Self-Selective Context. SSC scales sub-linearly with the number of fused contexts. And it leverages the correlation of human-object and context.

\section{Approach}

\subsection{Overview} 

The goal of this paper is to map an input image to the correct interaction class $I \rightarrow Y$.
If the image contains one pair of human and object, we can represent this pair as the feature $x_{ho}$ using off-the-shelf CNN~\cite{vgg}.
Then, using a classifier $C(\cdot)$, one can predict the probability scores of interaction classes $Y$.
However, this paper argues that recognizing the interaction based on only the the human-object feature $x_{ho}$ is sub-optimal.
It is preferred to complement $x_{ho}$ with more prior representations.
Several sources of contexts are the perfect choices for such prior representations.
As such, this paper contributes to the following. 1) We define new sources of contexts, along with their feature representations $\textbf{X}_c$, see section~\ref{sec:context}.
2) We propose a new method, Self-Selection Context (SSC) that learns how to complement the human-object feature $x_{ho}$ with the corresponding context features $\textbf{X}_c$, see section~\ref{sec:gating}.

\subsection{Self-Selection Context}
\label{sec:gating}

Given an input image $I$ comprising a set of $N$ human-object pairs.
Each pair possibly describes the interaction in such an image.
These $N$ pairs are represented as features $\mathbf{X}_{ho} = \{x_{ho}^{j}\}_{j=1}^{N}$  using off-the-shelf human-object extractor $f_{ho}(\cdot)$.
In addition, we are given a set of $M$ sources of contexts, corresponding to the input image $I$.
These contexts can be represented as features $\mathbf{X}_c = \{x_c^{i}\}_{i=1}^{M}$.
Each context feature $x_c^{i}$ is obtained from a different off-the-shelf context extractor $f_{c}^{i}(\cdot)$.
All of our feature extractors $[f_{ho}(\cdot), f_{c}^{i}(\cdot)]$ build upon the same CNN~\cite{vgg} for a fair comparison.
Hereafter, the \textit{global layer} is used to refer to the last fully convolutional layer \texttt{conv5\_3} of the CNN.

The goal of SSC, see Figure~\ref{fig:gating}\textcolor{red}{b}, is to complement each human-object feature $x^{j}_{ho}$ with the corresponding context features $\textbf{X}_c$.
That is why the core of the SSC is a novel Gating module $G(\cdot)$, see Figure~\ref{fig:gating}\textcolor{red}{a}.
$G(\cdot)$ serves two purposes.
First, as the context features are heterogeneous, comes from different extractors, and have different feature dimensions, $G(\cdot)$ embeds $\mathbf{X}_c$ into a compact features $\mathbf{Z}_c$ with a unified feature dimension.
Second, it predicts the gating vectors $\mathbf{\alpha} = \{ \alpha^{j} \}_{j=1}^{N}$, where $\alpha^{j} \in \Ree^{M}$ is the gating vector corresponding to the $j$-th human-object feature and all the $M$ context features $\mathbf{X}_c$.
After the Gating module, SSC complement each human-object features $x^{j}_{ho}$ with multitude of corresponding context features $\mathbf{Z}_{ho}$ in an adaptive manner:
\begin{equation}
 x^{j} = \texttt{cat} \left( x_{ho}^{j} \;,\; \sum_{i=1}^{M} \alpha^{ij}*z_{c}^{i}\right)\label{equoverview},
\end{equation}

\noindent where $\texttt{cat}(\cdot)$ is the concatenation operation along the feature dimension.
The output interaction feature $x^{j}$ is feed-forwarded to a Multi-Layer Perception (MLP) $C(\cdot)$ for interaction classification.


\subsection{Gating Module}

The main goal of the Gating module is to select (\textit{i.e.} gate) the most relevant context sources for each human-object pair $G(\cdot)$.
This gating helps in recognizing human interaction by incorporating the prior knowledge in such contexts.
Tthe gating mechanism $G(\cdot)$ is conditioned on the human-object $x_{ho}^{i}$ and it contexts $X_c$, \textit{jointly}.

The Gating module $G(\cdot)$ takes as an input the set of $N$ human-object features $\mathbf{X}_{ho} = \{x^{j}_{ho}\}_{j=1}^{N}$.
Also, $G(\cdot)$ takes the set of $M$ context features $\mathbf{X}_c = \{x^{i}_{c}\}_{i=1}^{M}$.
These features $\mathbf{X}_c$ are heterogeneous, each $x^{i}_{c}$ is obtained from a different context extractor $f_{c}^{i}(\cdot)$ with a different space dimension.
Thus, the first step is to embed each $x^{i}_{c}$ into a common dimension using linear mapping $g_{\psi}^{i}(\cdot)$.
The outcome is the embedded context features $\mathbf{Z}_{c} = \{z^{i}_{c}\}_{i=1}^{M}$.
Then, we need to measure the correlation between the human-object features $\mathbf{X}_{ho}$ and their corresponding context features $\mathbf{X}_{c}$.
But since both $\mathbf{X}_{ho}$ and $\mathbf{X}_{c}$ have different space dimensions, we use two linear embeddings $g_{\theta}(\cdot)$ and $g_{\phi}(\cdot)$ to map $\mathbf{X}_{ho}$ and $\mathbf{X}_{c}$, respectively, into two spaces with common dimension $C^{\prime}$,
\begin{align}
\mathbf{Q} &= g_{\theta}(\mathbf{X}_{ho})
\\
\mathbf{K} &= g_{\phi}(\mathbf{X}_c).
\end{align}
The outcome is the features $\textbf{Q} \in \Ree^{N \times C^{\prime}}$ and $\textbf{K} \in \Ree^{M \times C^{\prime}}$, respectively.
Then, we measure the pairwise correlation between the $N$ human-object pairs $\textbf{Q}$ and $M$ contexts $\textbf{K}$ using an inner product
\begin{align}
\alpha &=  \texttt{softmax}\left( \mathbf{Q} \odot \mathbf{K}^{\top} \right),
\end{align}
where $\alpha \in \Ree^{N \times M}, \mathbf{\alpha} = \{ \alpha^{j} \}^{N}_{j=1}$ are the gating vectors.
Each gating vector $\alpha^{j} \in \Ree^{M}$ represents how the $j$-th human-object pair correlated with all the $M$ contexts.
Notice that $\alpha$ is activated with \texttt{softmax} along with the $M$ contexts as a way of normalization.
%
%
The next step is to use the gating vector $\alpha^{j}$ to pool the embedded context features $\mathbf{Z_{c}}$ into the final context feature $x_{c}^{\prime} \in \Ree^{C^{\prime}}$.
$x_{c}^{\prime}$ is calculated as the inner product $\odot$ between $\alpha$ and $\mathbf{Z}_c$
\begin{align}
    x_{c}^{\prime} &= \alpha^{\top} \odot \mathbf{Z}_c.
\end{align}
To obtain the final interaction feature $x^j$, both the human-object feature $x_{ho}^j$ and its corresponding pooled context feature $x_{c}^{\prime}$ are concatenated along the feature dimension as shown in Equation~\ref{equoverview}.
This feature $x^j$ is feed-forwarded to the classifier $C(\cdot)$ to obtain the probability scores $Y = C(x^j)$ of classifying the interaction represented by input human-object pair $x^{j}_{ho}$.

\ptspace\partitle{Classifying interaction.} So far we have described how to combine human-object feature $x_{ho}^{j}$ with context features $\mathbf{X}_c$.
An image potentially has multiple human-objects ($N=12$ in our case), and it is not known which human-object pair is conducting the interaction (among all possible pairs). To that end, we resort to Multiple Instance Learning (MIL) as is the common practice~\cite{gkioxari2015contextual} to obtain the final image-level classifier response. Specifically, we first obtain the classifier predictions for each human-object pairings $Y \in \Ree^{N \times S}$ where $S$ denotes the number of interaction categories, then we apply max-pooling over the human-object dimension to obtain the final image-level interaction response ${Y}^{\prime} = \texttt{pool} \left( Y \right), {Y}^{\prime} \in \Ree^{S}$.

Worth mentioning that the classifier $C(\cdot)$ is a Multi-Layer Perceptron (MLP) with two hidden layers.
Each hidden layer is followed by  \texttt{BatchNorm} and \texttt{ReLU} non-linearity.
The output layer uses \texttt{sigmoid} non-linearity, to handle multiple labels per-image.
Up till now, we have described our method, SSC.
In the following, we first describe the human-object extractor $f_{ho}(\cdot)$.
Then, we complement with our contextual extractors $f_{c}^{i}(\cdot)$.


\begin{figure*}[!ht]
\begin{center}
\includegraphics[width=1.\textwidth]{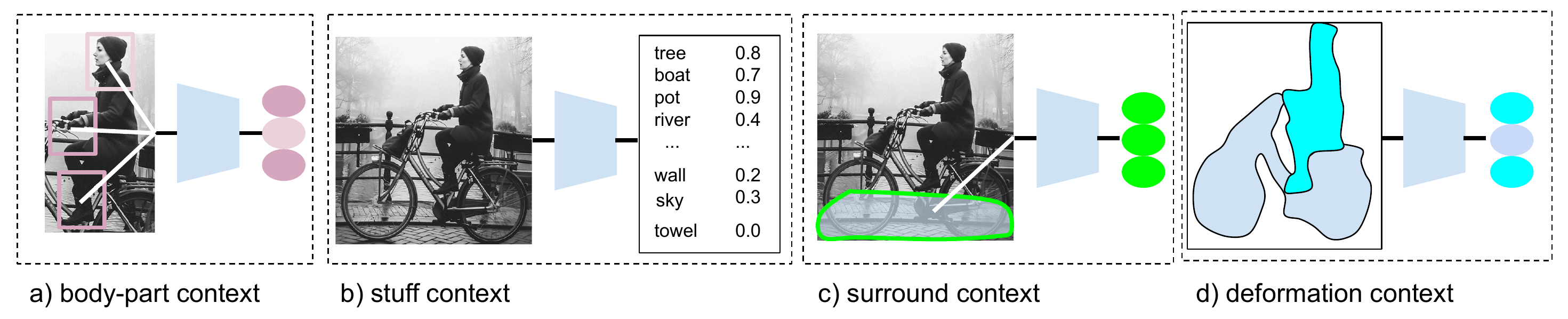}
\end{center}



\caption{The context features proposed in this paper. \texttt{a)} Body-part context models the appearance of the human joints, \texttt{b)} Stuff-context models the occurrence of stuff-like regions in the image, \texttt{c)} Surround context models the appearance of local segments around humans, \texttt{d)} Deformation context models the shape of the human-object posture.}

\label{fig:source}
\end{figure*}

\subsection{Human-object Features $\mathbf{X}_{ho}$}
\label{sec:basefeature}

Our human-object features are obtained from CNN~\cite{vgg} that takes as input an image and returns the global appearance features for $N$ human-object pairs in the image. To achieve this, we first detect possible human-object locations using an off-the-shelf object detector~\cite{faster-rcnn}. 
We select the top-$3$ detected humans and the top-$4$ detected objects, based on the detector confidence.
The permutations of $3$ humans and $4$ objects yield $3\times4=12$ distinct human-object pairs. 
We compute the union region for each pair, and then apply Region-of-Interest (ROI) pooling~\cite{fastrcnn} over this region to obtain the final features per human-object.
ROI pooling is applied over the global layer of the network.

\subsection{Contextual Features $\mathbf{X}_{c}$}
\label{sec:context}


For an image, we extract $M$ different context features, $\mathbf{X}_c$, see Figure~\ref{fig:source}, where $M=4$ where each feature $x_c^{i}$ is obtained from a different extractor $f_c^{i}(\cdot)$, which we detail below.

\partitle{Body-part context feature.} This context, see Figure~\ref{fig:source}\textcolor{red}{a}, models the local body-parts of the human interactor. Local body-parts carry distinctive information about an interaction, especially for grasping. The context is implemented as follows.

First, the context requires an initial set of human body-part regions. Hence, given each detected human, we run an off-the-shelf human keypoint detector~\cite{alphapose}. This yields $17$ distinct body-part regions such as the knee, the hand or the head. We draw a regular bounding box around each keypoint region. Given these bounding boxes, we apply ROI pooling over each region for each detected human from the global layer of the CNN. 

However, not all body-parts contribute equally to the interaction. To select the most discriminative body-part, we then feedforward each ROI-pooled part feature to an attentional sub-network $f_{att}(\cdot)$ that yields a scalar value per-part indicating their relevance. Based on the obtained scores, we only select top-$k$ regions for further processing ($k=3$). Nonetheless, this indexing operation is non-differentiable. To that end, we employ a strategy called straight-through estimator~\cite{straight} to by-pass the gradient computation for the non-selected body-parts. In practice, the gradients of the non-selected body-parts are set to $0$. Finally, the selected body-part features are concatenated and further compressed with two fully connected layers, leading to a $1k$ dimensional feature summarizing distinctive human parts.

\partitle{Stuff context feature.} This context models the existence of local stuff-like regions, such as trees, wall, river, see Figure~\ref{fig:source}\textcolor{red}{b}. The existence of such regions can give hint about the interactions, \eg river for boat riding.

We implement the stuff-context as the class probabilities of a stuff-classifier. Stuff-classifier is a linear classifier on top of the global layer response of the CNN using the annotations from~\cite{cocostuff}. The final response is a $91-$dimensional feature vector summarizing the existence of local stuff-like regions in the image.

\partitle{Surround context feature.} This context models the local surround around human-object, see Figure~\ref{fig:source}\textcolor{red}{c}. To represent the local surround, we make use of the semantic segments like the road, the sky or the sideways, obtained from an off-the-shelf model~\cite{zhou2019semantic} over the input image. 

Specifically, we first create binary masks from the input image of size $m \in \Ree^{H\times W\times K}$, where $[H,W]$ denotes the height and width of the global layer feature map and $K$ is the number of distinct segments. For each image, we choose the top $K$ segments with the largest scales ($K=5$). For each mask, the values of only the respective segment are set to $1$ and all else is set to $0$. Masks are then applied to the global activations (conv$5\_3$) of the CNN, which is then average-pooled over the spatial dimension to obtain the features of size $\Ree^{K \times D}$. Finally, we aggregate over the semantic segment dimension $K$ to obtain $\Ree^{1 \times D}$ feature response ($d=512$) via max pooling.

\partitle{Deformation context feature.} This context models the mutual deformation of human-objects, see Figure~\ref{fig:source}\textcolor{red}{d}. Our goal is to encode two important cues of interactions: 1) Shape of the human-object deformation, 2) Spatial relation of the human-object location simultaneously. 

To that end, given an image, we first obtain object segmentation predictions~\cite{maskrcnn} of size $\Ree^{H \times W \times K}$ where $[H,W]$ are the heights and width of the image, and $K$ is the number of distinct objects ($K=80$). We resize this feature map such that the longest side is $64$ pixels, using bi-linear interpolation, and process it with a three-layer CNN along the channel dimensions $80\rightarrow128\rightarrow256\rightarrow512$. The first layer is a $1\times1$ convolution followed by a $3\times3$ kernel. 
Keeping the first convolution $1\times 1$ is crucial -- The input mask is sparse (\ie most locations are $0$), which is hard to process with a dense filter of size $3\times3$ from the start. To that end, we first generate a denser feature map using $1\times1$ filters to process with subsequent layers.
Finally, we pool the response over the spatial dimension to obtain $512$-dimensional deformation context feature. 

\partitle{Implementation details.} All the models are implemented in PyTorch~\cite{pytorch}, trained and optimized with SGD. The CNN~\cite{vgg} is pre-trained on ImageNet~\cite{imagenet}.
Then it is fine-tuned on the HICO dataset for $epochs$ with a learning rate $0.001$ that is decayed by a factor $0.1$ after $15$ epochs.

\section{Datasets and Setup}

\subsection{Contextualized Interactions Dataset}

Existing datasets~\cite{hico,vcoco} are limited in their context repertoire since they are collected with \texttt{<verb, noun>} queries only. 
This prevents observing the contribution of the context in diverse environments. To that end, we collect a new challenging dataset we name \textbf{C}ontextualized \textbf{Int}eractions (CINT).

To create CINT, we first queried Google Images~\cite{google} via triplets of \texttt{<verb, noun, context>} queries, where \texttt{<verb, noun>} pairs are from HICO and $40$ context queries are derived from scene datasets~\cite{scene-sun-database,scene-places,scene-attribute}. Context queries are the time of the day (\eg day, night, $7/40$), state of the scene (\eg sunrise, snowy, dark, $12/40$) or the location of the scene (\eg railroad, beach, street, $21/40$). After omitting the queries with no results, we ended up with around $2k$ distinct \texttt{<verb, noun, context>} queries of which we use in the annotation procedure. 

We initially downloaded $250$ images per-query and then removed images where the human, the object and the context are not visible. We have $16k$ images from $200$ distinct interactions. 

\subsection{Other Datasets}

\partitle{HICO dataset~\cite{hico}.} For training we rely on HICO. It contains $40k$ training and $10k$ testing images from $600$ distinct \texttt{<verb, noun>} pairs, for $117$ verbs and $80$ nouns. HICO exhibits a long-tailed distribution: $155$ classes have less than $10$ examples. A special case is \texttt{no-interaction}, where the target object and a human is visible whereas not interacting.  

\partitle{V-COCO dataset~\cite{vcoco}.} V-COCO was initially designed for interaction detection~\cite{chao2018learning,gkioxari2018detecting}. 
To demonstrate the generality of our approach, we re-purpose the dataset for interaction recognition as follows.
We align class namings with HICO, for example by splitting  \texttt{<skateboarding>} into \texttt{<ride, skateboard>}). We omit actions like smiling that do not correspond to any object. 
Finally, we follow the best practice of~\cite{hico} and aggregate different interaction instances within the same image over the image label.
As a result, V-COCO has $4k$ test images with $226$ distinct interactions. A big portion of the dataset belongs to interactions with rare categories, making the inference challenging.

\subsection{Baseline models}

\partitle{Interaction recognition baselines.} To prove the generality of our approach with different human-object representations, we plug in our SSC to three different human-object feature extractors, namely: 1) VGG-$16$~\cite{vgg} that is pre-trained on ImageNet and fine-tuned on HICO, 2) ContextFusion~\cite{mallya2016learning}, 3) Global stream of PairAtt~\cite{intrec-partattention}. All the features are extracted from the penultimate layer of \texttt{fc7} using the code from the respective authors.



\partitle{Fusion baseline.} We compare SSC to a fusion baseline, where we concatenate human-object features with different context features. 

\subsection{Evaluation}
For evaluation, we use the instance-based Mean Average Precision (mAP). That is, we evaluate the prediction performance per-image, which is then averaged over the respective dataset. This metric allows us to observe the effect of fusing or Self-Selecting different forms of context features on a per-image basis. Additionally, it allows us to analyze the effect of the context over the characteristics of interactions such as the size, the population, or the existence of interaction.

\section{Experiments}



\subsection{Self-Selection of the Single Context}



In the first experiment, we validate the discriminative ability of the proposed context features on HICO. We select and combine single context features with the human-object feature, to see their individual contributions. Results are in Table~\ref{tab:exp1}.

\begin{table}[h]
   \caption{Self-Selection of The Single Context.}

\begin{center}
\setlength\tabcolsep{12.5pt}
\begin{tabular}{lcc}

\toprule
 Context Feature  & mAP(\%) & Improvement $\Delta \uparrow$ \\
\midrule 

 Human-Object (HO)-only  & $62.50$ & - \\
\midrule 

   HO + Body-part context & $68.55$ & $6.05$ \\ 

\midrule
 HO + Stuff context & $68.20$ & $5.70$ \\

\midrule

HO + Surround context & $68.44$ & $5.94$ \\

\midrule

 HO + Deformation context &  $68.30$ & $5.80$ \\

\midrule

\end{tabular}
\label{tab:exp1}
\end{center}

\end{table}

Each of the $4$ context features, by itself, are complementary to the human-object feature. Each improves the performance of human-object features considerably.

Body-part context helps the most with $6.05$ mAP since many interactions are localized on the fine-grained body-parts such as the hand-object interactions like cutting, eating or cooking. Also, the surround context helps with $5.94$ mAP, indicating that the immediate surround of the human-objects is distinctive, such as the road for transport vehicles. Deformation context helps with $5.80$ mAP since it complements the human-object features with the mutual position and the deformation information, which is crucial in dynamic interactions like jumping or throwing. Lastly, stuff context helps with $5.70$ mAP which confirms that high-level representation of surrounding object-like regions can help distinguish the interaction. 


To conclude, we observe that context features are distinctive for recognition and complement human-object features. Also, each different context feature specializes in different interactions which call for an effective combination. 

\subsection{Self-Selection of the Multiple Contexts}

This experiment validates the complementary power of the proposed context features via Self-Selection. We select and combine all $4$ contextual features with human-objects. Results are in Table~\ref{tab:exp2}.


   



\begin{table}[h]
   \caption{Self-Selection of The Multiple Contexts.}
\begin{center}
\renewcommand{\arraystretch}{1.0}
\setlength\tabcolsep{14.0pt}
\begin{tabular}{lccc}

\toprule 
&   \multicolumn{3}{c}{Dataset} \\
\cmidrule(l r){2-4}

Method  & HICO & V-COCO & CINT \\ 
\midrule 
HO-only  & $62.50$  & $52.27$   & $45.24$ \\
\midrule 
HO + Fusion  & $69.59$  & $54.74$   & $49.74$ \\
 \midrule 
HO + SSC (Ours)  & $\textbf{70.78}$  & $\textbf{55.00}$   & $\textbf{54.36}$ \\

\bottomrule
\end{tabular}

\label{tab:exp2}
\end{center}
\end{table}








It is observed that both fusion and SSC improves upon human-object-only feature, confirming the complementary power of the proposed contexts. We also see that SSC performs better than fusion across all three datasets. The difference is more pronounced on CINT, which highlights the need for selecting the discriminative context in diverse environments.

\begin{figure}[h]
\begin{center}
\includegraphics[width=0.9\linewidth]{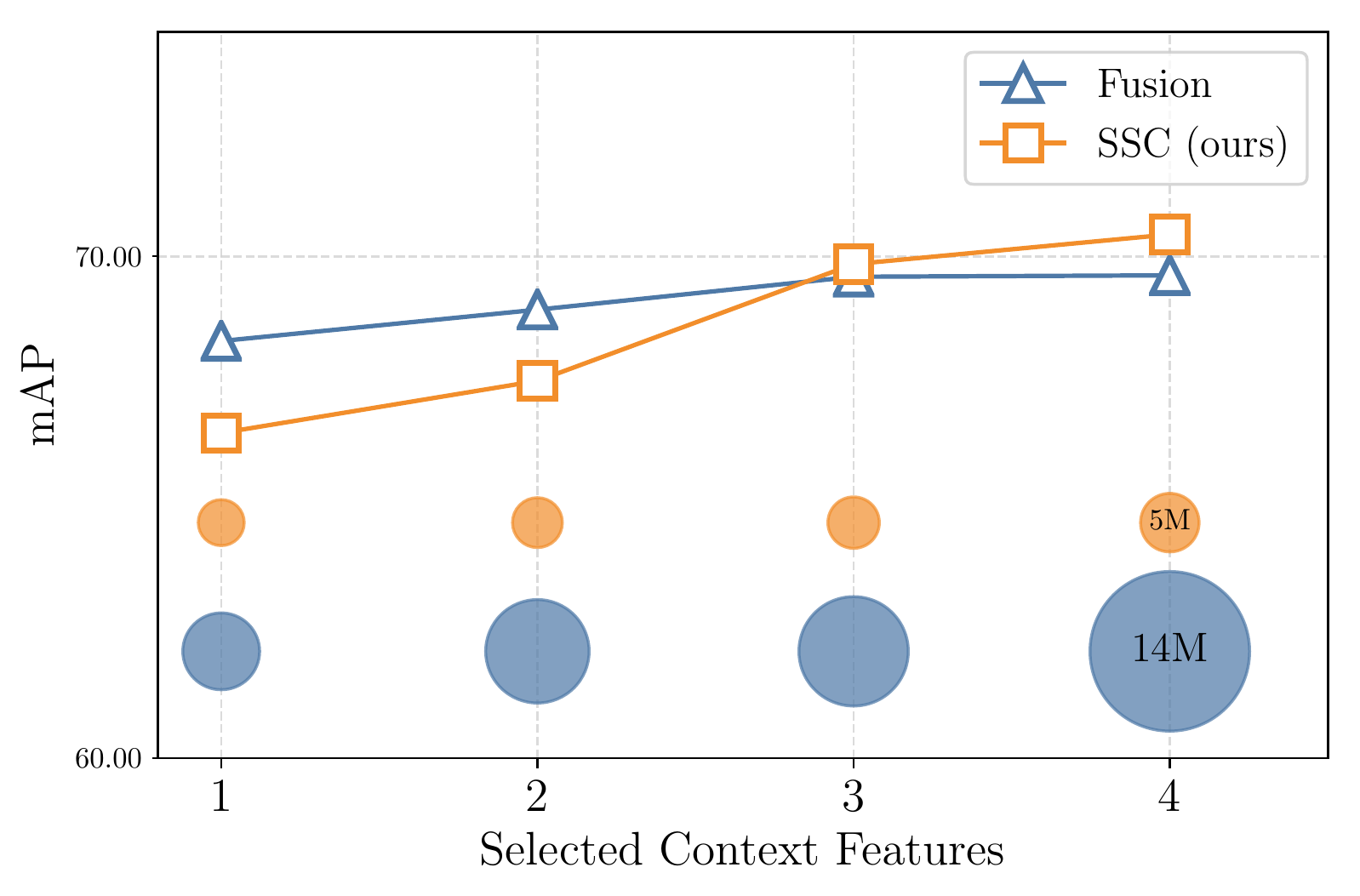}
\end{center}


\caption{Parameter efficiency of Self-Selection vs. Fusion. Amount of parameters for each step for the respective technique is presented in log-scale using circles.}

\label{fig:param-eff}
\end{figure}

SSC does so by using three times fewer parameters, as can be seen from Figure~\ref{fig:param-eff}. Figure~\ref{fig:param-eff} plots the recognition mAP as a function of the number of cumulative contexts added at each step (from $1$ to $4$). We initially add stuff context and then add $1$ more context at each step. As can be seen, Self-Selective Context uses $3$ times fewer parameters than the fusion counterpart ($4.9$ Million vs. $13.6$ Million) while yielding better results. This is expected since many human-object interactions have limited examples in the training set, hence making the learning difficult. 

To conclude, we observe that the context features are complementary to each other, and, Self-Selective context provides a parameter efficient and accurate combination of contexts.

\subsection{Further Analysis of Self-Selection}

In this Section, we present further analysis of Self-Selection, on the source of the improvement, the ablation of the joint conditioning, and the distribution of Self-Selection. 


\partitle{Source of the improvement.} To shed some light into where the gain comes from, we marginalize the improvement of SSC over human-object features in Table~\ref{tab:secondorder}. 

\begin{table}[h]

   \caption{The Source of the Improvement.}

\resizebox{\columnwidth}{!}{%
\begin{tabular}{ccccccc}

\toprule                       
 \multirow{2}{*}{Method}   & \multicolumn{2}{c}{Pixel Area}  
         & \multicolumn{2}{c}{Population} 
         & \multicolumn{2}{c}{Existence}
\\
\cmidrule(l r){2-3} \cmidrule(l r){4-5} \cmidrule(l r){6-7}

                & Small & Large & Rare & Frequent & No & Yes  \\        
\midrule

HO-only & $60.09$ & $63.98$ & $39.31$  & $61.68$  & $36.96$ & $64.26$   \\
HO + \textbf{SSC} & $69.45$  & $72.70$ & $51.38$  & $70.79$  & $47.67$ & $73.24$  \\
\midrule
$\Delta \uparrow$ & $\textbf{9.36}$ & $8.72$ & $\textbf{12.07}$ & $9.11$ & $\textbf{10.71}$ & $8.98$ \\  

\bottomrule
\end{tabular}
}

\label{tab:secondorder}
\end{table}

1) SSC helps slightly better for small human-object interactions. During this experiment, an interaction is deemed to be small if the human-objects occupy less than $20\%$ of the whole image, and large otherwise. This indicates that when the visual details of the human-objects are limited due to size, the context becomes more important. 

2) SSC helps considerably better for rare human-object interactions. During this experiment, an interaction is deemed to be rare if it has less than $10$ examples in the HICO training set, frequent otherwise. This indicates that rare human-object interactions (\ie \texttt{<ride, giraffe>}) exhibit distinctive local contextual appearance that, once modeled with SSC, becomes easier to recognize. 

3) SSC helps considerably better for the case of no-interaction. During this experiment, we aggregate the performance over no-interaction and interaction categories separately. This indicates that SSC encodes the distinctive signals of the interaction, which is once used, helps the model to discriminative interaction from no-interaction.

\partitle{Contribution of the joint conditioning.} For an ablation, study we remove the joint conditioning from SSC. In this way, our model learns context relevance values $\alpha$ by considering the \textit{context only} (as opposed to human-object and context together). Results can be seen from Table~\ref{tab:cond}. 

\begin{table}[h]

\caption{Contribution of Joint Conditioning.}

\begin{center}
    
\setlength\tabcolsep{14.0pt}

\begin{tabular}{cccc}
\toprule  
  
  Condition  & \multicolumn{3}{c}{Dataset} \\ 
 
   \cmidrule(l r){2-4}
   
 & HICO & V-COCO & CINT \\
   
\midrule



  context-\textit{only} & 67.77          & 49.58          & 49.26  \\
  human-object \& context  & \textbf{70.78} & \textbf{55.00} & \textbf{54.36}  \\

\bottomrule
\end{tabular}

\label{tab:cond}
\end{center}
\end{table}


As can be seen, using context-\textit{only}  to select the context leads to a drop over all three datasets, confirming the importance of Self-Selection jointly based on the human-object and context.

\begin{figure}[h]
\begin{minipage}[b]{0.9\linewidth}
\centering
\includegraphics[width=1\linewidth]{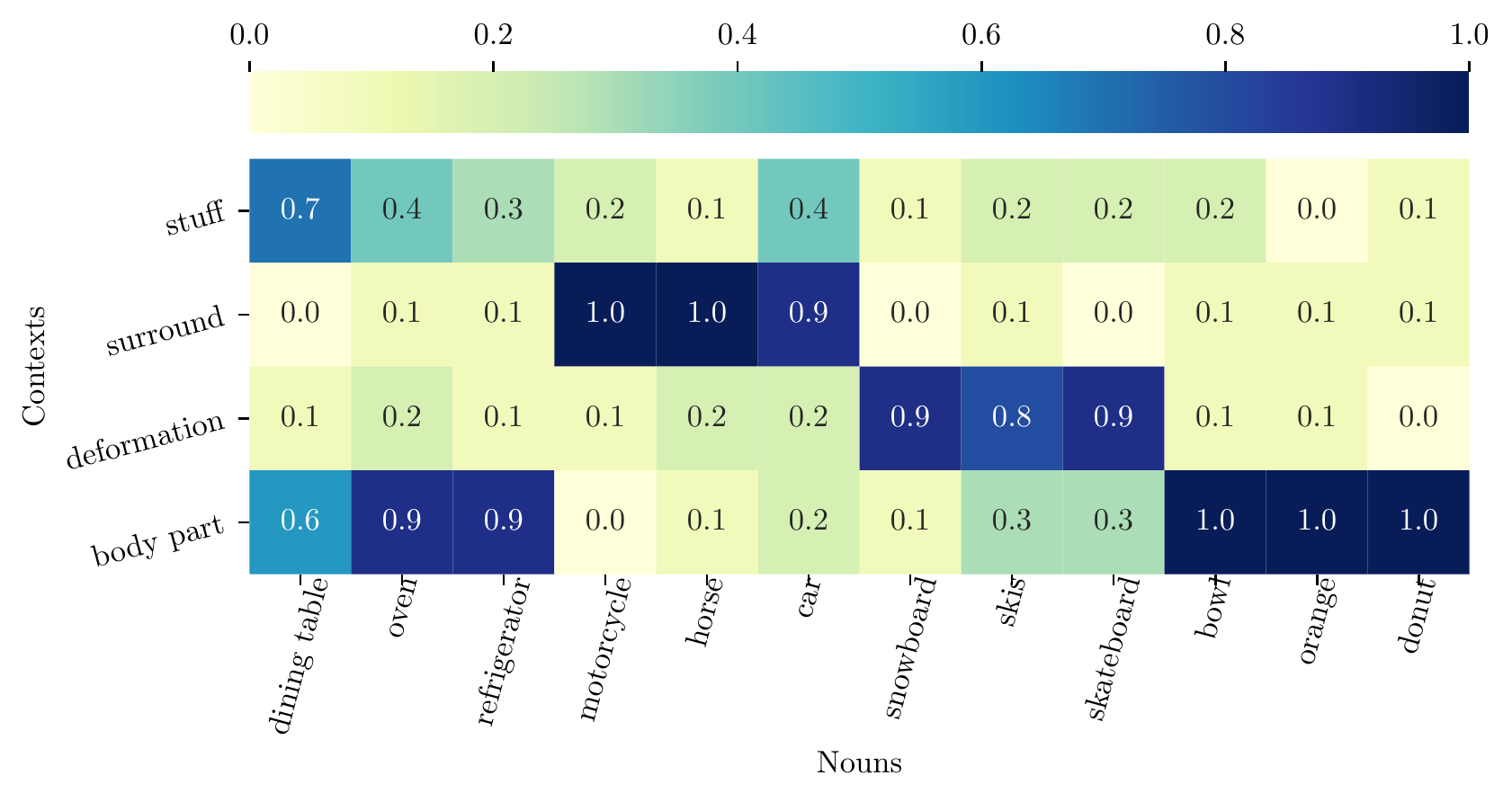}
\end{minipage}
\vspace{1mm}
\begin{minipage}[b]{0.9\linewidth}
\centering
\includegraphics[width=1\linewidth]{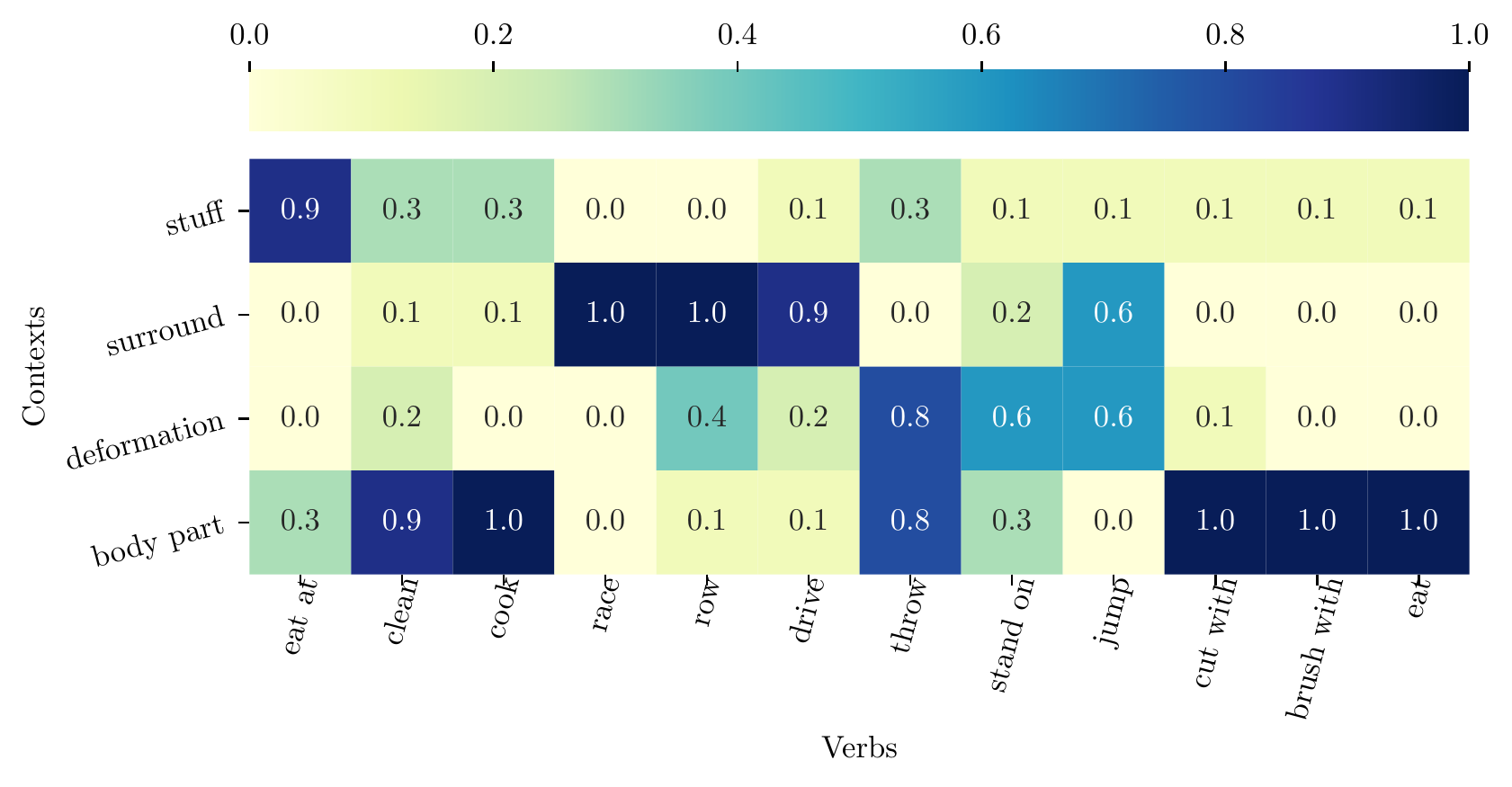}
\end{minipage}
\caption{Distribution of Self-Selection over nouns and verbs.}
\label{fig:contsel}
\end{figure}

\partitle{Distribution of Self-Selection.} In this experiment we visualize the distribution of Self-Selection. We aggregate Self-Selection ratios over distinct nouns and verbs in HICO in Figure~\ref{fig:contsel}.  

Stuff context is preferred in house activities of cooking and cleaning, where the co-occurring objects are distinctive. The surround context is preferred by transportation interactions that use horses, cars, or motorcycles for racing or driving. Deformation context is preferred by sport objects like skateboard or skis for jumping or standing, where the interaction leads to unique postures in the human-object body. Lastly, hand-object interactions like cutting, brushing, or eating prefer body-part context, where the hand leads to distinctive occlusion patterns over the object region. We re-assure that the contribution of the Self-Selection is based on the interaction type. 




\subsection{Self-Selection with the State-of-the-art}

So far, we experimented with our proposed human-object features. This experiment plugs in our Self-Selective module to existing models, namely VGG-$16$~\cite{vgg}, ContextFusion~\cite{mallya2016learning}, and global stream of PairAtt~\cite{intrec-partattention}. Results from the three datasets can be seen from Table~\ref{tab:exp3}.

\begin{table}[h]

\caption{Combining SSC with the State-of-the-art.}

\begin{center}
    
\setlength\tabcolsep{14.0pt}
\renewcommand{\arraystretch}{1.5}

\begin{tabular}{lccc}
\toprule  
  \multirow{2}{*}{\hspace{30pt}Method}  & \multicolumn{3}{c}{Dataset} \\ 
 
   \cmidrule(l r){2-4}
   
 & HICO & V-COCO & CINT \\
   
\midrule



 VGG-16~\cite{vgg}  & 56.10 & 46.91 & 44.83  \\
   VGG-16~\cite{vgg} + \textbf{SSC} & \textbf{67.59} & \textbf{51.17} & \textbf{49.56}  \\
\hline 

ContFus~\cite{mallya2016learning} & 63.47 & 51.36 & 46.72  \\
  ContFus~\cite{mallya2016learning} + \textbf{SSC} & \textbf{65.79} & \textbf{52.24} & \textbf{51.92}  \\
\hline 

PairAtt~\cite{intrec-partattention}  & 65.10 & 53.62 & 48.99  \\
  PairAtt~\cite{intrec-partattention} + \textbf{SSC} & \textbf{68.29} & \textbf{54.24} & \textbf{51.22}  \\
 
 \hline

Human-Object & 62.50 & 51.60 & 47.85  \\
 Human-Object + \textbf{SSC} & \textbf{70.78} & \textbf{55.00} & \textbf{54.36}  \\

\bottomrule
\end{tabular}

\label{tab:exp3}
\end{center}
\end{table}

As can be seen in all cases, incorporating Self-Selective context improves upon the respective model alone. This indicates that Self-Selective context carries complementary information for State-of-the-art models, even though these features incorporate some contexts like global surround or object co-occurrence intrinsically. An important result is that human-object features coupled with Self-Selective context still outperforms all other models, despite its simplicity. This indicates that modeling human-object and context separately and adaptively is essential for recognizing human-object interactions.

\begin{figure}[h]
\begin{center}
\includegraphics[width=1.0\linewidth]{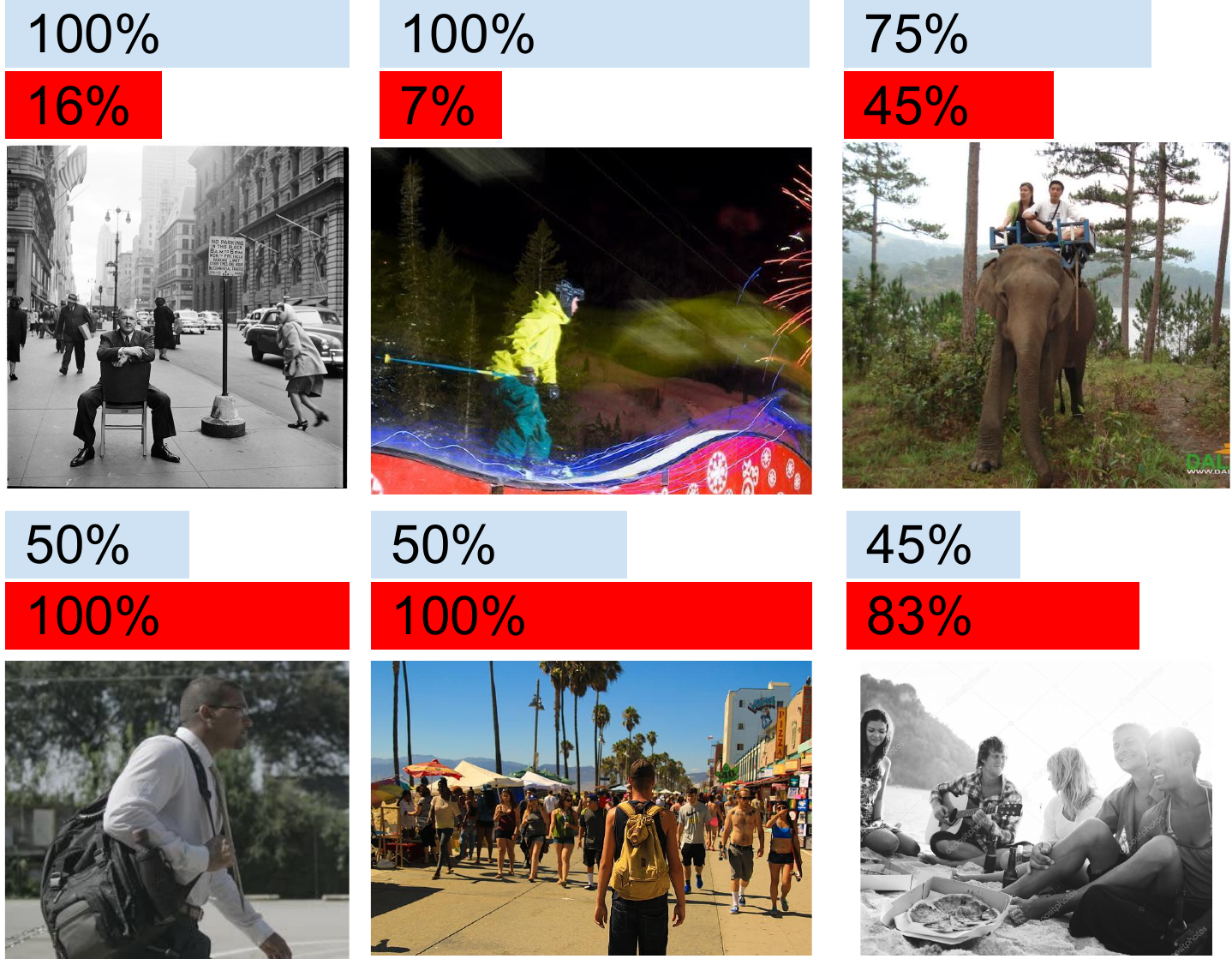}
\end{center}
\caption{Qualitative examples from CINT dataset for \textcolor{red}{PairAtt}~\cite{intrec-partattention} and \textcolor{blue}{PairAtt+SSC(ours)}. We provide mAP $\%$ on top for both. SSC helps when the context is unexpected (top), however may decrease the result if the context is not visible or too noisy (bottom).}
\label{fig:qualitative}
\end{figure}

\partitle{Qualitative analysis.} We present success and failure cases in Figure~\ref{fig:qualitative}.
We compare the performance of PairAtt with PairAtt + \textbf{SSC} using images from CINT dataset. 

In the top, we provide three examples where SSC improves upon PairAtt. We can see that SSC helps when the scene is not strongly correlated with the target interaction, such as \texttt{<sit on, chair>} on city street, or \texttt{<ride, ski>} in the night. In such particular cases, SSC can suppress the contribution of the irrelevant context feature, therefore leading to accurate classification performance. 

In the bottom, we provide three examples SSC decreases the performance. We can see that when the amount of visual context is limited, as in \texttt{<carry, backpack>} on left bottom example, or when the context is too noisy, such as the background humans in the middle bottom, SSC is challenged in identifying the discriminative context. This leaves for improvement for such cases. 

To conclude, Self-Selective context carries information for the State-of-the-art models, as shown in Table~\ref{tab:exp3}. The difference is considerable for all three models, on HICO, V-COCO, and CINT. Also, Self-Selective context helps the most when the context is radically different as shown in Figure~\ref{fig:qualitative}.

\section{Conclusion}

In this work, we addressed the task of recognizing human-object interactions from a single image. We treated human-object interaction recognition as a task of context selection. We first devised context features to model the locality of the human, the object, the surround scene, and the human-objects. Then, we proposed a new model, namely scalable Self-Selective Context (SSC), along with a novel gating module.
The gating mechanism considers the correlation between the human-objects and the corresponding contexts.
And the gating module succeeds in selecting the context(s) that are most relevant to the interactions in each image.
Our experiments reveal that, indeed, the proposed context features are discriminative and they complement the appearance features of human-objects.
In addition, our method, SSC, improves State-of-the-art in interaction recognition on three challenging benchmarks: HICO, V-COCO and CINT.

{\small
\bibliographystyle{./IEEEtran}
\bibliography{egbib}

\begin{thebibliography}{10}
\providecommand{\url}[1]{#1}
\csname url@samestyle\endcsname
\providecommand{\newblock}{\relax}
\providecommand{\bibinfo}[2]{#2}
\providecommand{\BIBentrySTDinterwordspacing}{\spaceskip=0pt\relax}
\providecommand{\BIBentryALTinterwordstretchfactor}{4}
\providecommand{\BIBentryALTinterwordspacing}{\spaceskip=\fontdimen2\font plus
\BIBentryALTinterwordstretchfactor\fontdimen3\font minus
  \fontdimen4\font\relax}
\providecommand{\BIBforeignlanguage}[2]{{%
\expandafter\ifx\csname l@#1\endcsname\relax
\typeout{** WARNING: IEEEtran.bst: No hyphenation pattern has been}%
\typeout{** loaded for the language `#1'. Using the pattern for}%
\typeout{** the default language instead.}%
\else
\language=\csname l@#1\endcsname
\fi
#2}}
\providecommand{\BIBdecl}{\relax}
\BIBdecl

\bibitem{lallee2010human}
S.~Lall{\'e}e, E.~Yoshida, A.~Mallet, F.~Nori, L.~Natale, G.~Metta,
  F.~Warneken, and P.~F. Dominey, ``Human-robot cooperation based on
  interaction learning,'' in \emph{From motor learning to interaction learning
  in robots}, 2010.

\bibitem{yu2018one}
T.~Yu, C.~Finn, A.~Xie, S.~Dasari, T.~Zhang, P.~Abbeel, and S.~Levine,
  ``One-shot imitation from observing humans via domain-adaptive
  meta-learning,'' \emph{arXiv preprint}, 2018.

\bibitem{edsinger2007human}
A.~Edsinger and C.~C. Kemp, ``Human-robot interaction for cooperative
  manipulation: Handing objects to one another,'' in \emph{RO-MAN}, 2007.

\bibitem{herdade2019image}
S.~Herdade, A.~Kappeler, K.~Boakye, and J.~Soares, ``Image captioning:
  Transforming objects into words,'' in \emph{NeurIPS}, 2019.

\bibitem{mallya2016learning}
A.~Mallya and S.~Lazebnik, ``Learning models for actions and person-object
  interactions with transfer to question answering,'' in \emph{ECCV}, 2016.

\bibitem{gkioxari2015actions}
G.~Gkioxari, R.~Girshick, and J.~Malik, ``Actions and attributes from wholes
  and parts,'' in \emph{ICCV}, 2015.

\bibitem{intrec-partattention}
H.-S. Fang, J.~Cao, Y.-W. Tai, and C.~Lu, ``Pairwise body-part attention for
  recognizing human-object interactions,'' in \emph{ECCV}, 2018.

\bibitem{vaswani2017attention}
A.~Vaswani, N.~Shazeer, N.~Parmar, J.~Uszkoreit, L.~Jones, A.~N. Gomez,
  {\L}.~Kaiser, and I.~Polosukhin, ``Attention is all you need,'' in
  \emph{NeurIPS}, 2017.

\bibitem{chao2018learning}
Y.-W. Chao, Y.~Liu, X.~Liu, H.~Zeng, and J.~Deng, ``Learning to detect
  human-object interactions,'' in \emph{WACV}, 2018.

\bibitem{gkioxari2018detecting}
G.~Gkioxari, R.~Girshick, P.~Doll{\'a}r, and K.~He, ``Detecting and recognizing
  human-object interactions,'' in \emph{CVPR}, 2018.

\bibitem{hico}
Y.-W. Chao, Z.~Wang, Y.~He, J.~Wang, and J.~Deng, ``Hico: A benchmark for
  recognizing human-object interactions in images,'' in \emph{ICCV}, 2015.

\bibitem{vcoco}
S.~Gupta and J.~Malik, ``Visual semantic role labeling,'' \emph{arXiv
  preprint}, 2015.

\bibitem{fusionbagwords}
X.~Peng, L.~Wang, X.~Wang, and Y.~Qiao, ``Bag of visual words and fusion
  methods for action recognition: Comprehensive study and good practice,''
  \emph{CVIU}, 2016.

\bibitem{karpathy2014large}
A.~Karpathy, G.~Toderici, S.~Shetty, T.~Leung, R.~Sukthankar, and L.~Fei-Fei,
  ``Large-scale video classification with convolutional neural networks,'' in
  \emph{CVPR}, 2014.

\bibitem{fusionmotion}
K.~Simonyan and A.~Zisserman, ``Two-stream convolutional networks for action
  recognition in videos,'' in \emph{NeurIPS}, 2014.

\bibitem{vgg}
K.~Simonyan and A.~Ziserman, ``Very deep convolutional networks for large-scale
  image recognition,'' in \emph{ICLR}, 2015.

\bibitem{gkioxari2015contextual}
G.~Gkioxari, R.~Girshick, and J.~Malik, ``Contextual action recognition with r*
  cnn,'' in \emph{ICCV}, 2015.

\bibitem{faster-rcnn}
S.~Ren, K.~He, R.~Girshick, and J.~Sun, ``Faster r-cnn: Towards real-time
  object detection with region proposal networks,'' in \emph{NeurIPS}, 2015.

\bibitem{fastrcnn}
R.~Girshick, ``Fast r-cnn,'' in \emph{ICCV}, 2015.

\bibitem{alphapose}
H.-S. Fang, S.~Xie, Y.-W. Tai, and C.~Lu, ``{RMPE}: Regional multi-person pose
  estimation,'' in \emph{ICCV}, 2017.

\bibitem{straight}
Y.~Bengio, N.~L{\'e}onard, and A.~Courville, ``Estimating or propagating
  gradients through stochastic neurons for conditional computation,''
  \emph{arXiv preprint}, 2013.

\bibitem{cocostuff}
H.~Caesar, J.~Uijlings, and V.~Ferrari, ``Coco-stuff: Thing and stuff classes
  in context,'' in \emph{CVPR}, 2018.

\bibitem{zhou2019semantic}
B.~Zhou, H.~Zhao, X.~Puig, T.~Xiao, S.~Fidler, A.~Barriuso, and A.~Torralba,
  ``Semantic understanding of scenes through the ade20k dataset,'' \emph{IJCV},
  2019.

\bibitem{maskrcnn}
K.~He, G.~Gkioxari, P.~Doll{\'a}r, and R.~Girshick, ``Mask r-cnn,'' in
  \emph{ICCV}, 2017.

\bibitem{pytorch}
A.~Paszke, S.~Gross, S.~Chintala, G.~Chanan, E.~Yang, Z.~DeVito, Z.~Lin,
  A.~Desmaison, L.~Antiga, and A.~Lerer, ``Automatic differentiation in
  pytorch,'' 2017.

\bibitem{imagenet}
J.~Deng, W.~Dong, R.~Socher, L.-J. Li, K.~Li, and L.~Fei-Fei, ``Imagenet: A
  large-scale hierarchical image database,'' in \emph{CVPR}, 2009.

\bibitem{google}
Google, ``Google image search engine toolkit,''
  \url{https://developers.google.com/image-search/v1/devguide}, 2015.

\bibitem{scene-sun-database}
J.~Xiao, J.~Hays, K.~A. Ehinger, A.~Oliva, and A.~Torralba, ``Sun database:
  Large-scale scene recognition from abbey to zoo,'' in \emph{CVPR}, 2010.

\bibitem{scene-places}
B.~Zhou, A.~Lapedriza, A.~Khosla, A.~Oliva, and A.~Torralba, ``Places: A 10
  million image database for scene recognition,'' \emph{TPAMI}, 2018.

\bibitem{scene-attribute}
P.-Y. Laffont, Z.~Ren, X.~Tao, C.~Qian, and J.~Hays, ``Transient attributes for
  high-level understanding and editing of outdoor scenes,'' \emph{TOG}, 2014.

\end{thebibliography}
}

\end{document}